\documentclass[acmsmall,nonacm]{acmart}

\AtBeginDocument{%
  }

\renewcommand\footnotetextcopyrightpermission[1]{} % no permission footnote
\usepackage{xcolor}
\usepackage[ruled,linesnumbered,lined]{algorithm2e}
\usepackage{algpseudocode}
\usepackage{caption,subcaption}
\usepackage{float}
\usepackage{bm}

\newtheorem{thm}{Theorem}
\newtheorem{crl}{Corollary}
\newtheorem{lm}{Lemma}

\newcommand{\bx}{\mathbf{x}}

\begin{document}

\title{Bayesian Optimisation: Which Constraints Matter?}

\author{Xietao Wang Lin}
\affiliation{%
  \institution{Mathematics for Real-World Systems, University of Warwick}
  \city{Coventry}
  \country{UK}
}
\email{xietao.wang-lin@warwick.ac.uk}

\author{Juan Ungredda}
\affiliation{%
  \institution{ESTECO SpA}
  \city{Trieste}
  \country{Italy}
}
\email{ungredda@esteco.com}

\author{Max Butler}
\affiliation{%
  \institution{Mathematics for Real-World Systems, University of Warwick}
  \city{Coventry}
  \country{UK}
}
\email{max.butler@warwick.ac.uk}

\author{James Town}
\affiliation{%
  \institution{Mathematics for Real-World Systems, University of Warwick}
  \city{Coventry}
  \country{UK}
}
\email{james.p.town@warwick.ac.uk}

\author{Alma Rahat}
\affiliation{%
  \institution{Swansea University}
  \city{Swansea}
  \country{UK}
}
\email{a.a.m.rahat@swansea.ac.uk}

\author{Hemant Singh}
\affiliation{%
  \institution{University of New South Wales}
  \city{Canberra}
  \country{Australia}
}
\email{hemant.singh@unsw.edu.au}

\author{Juergen Branke}
\affiliation{%
  \institution{Warwick Business School, University of Warwick}
  \city{Coventry}
  \country{UK}
}
\email{juergen.branke@wbs.ac.uk}

\renewcommand{\shortauthors}{Lin et al.}

\begin{abstract}
Bayesian optimisation has proven to be a powerful tool for expensive global black-box optimisation problems. In this paper, we propose new Bayesian optimisation variants of the popular Knowledge Gradient acquisition functions for problems with \emph{decoupled} black-box constraints, in which subsets of the objective and constraint functions may be evaluated independently. In particular, our methods aim to take into account that often only a handful of the constraints may be binding at the optimum, and hence we should evaluate only relevant constraints when trying to optimise a function. 
%All of our methods are in the Bayesian Optimization paradigm, which relies on the assumption that all function evaluations are expensive, and hence we have only a very limited budget of evaluations. 
%We propose two algorithms which extend two existing coupled approaches, and call them  `decoupled constrained Knowledge Gradient' and `decoupled Expected Improvement', and we further propose a small modification to the latter as a separate algorithm, which we call hybrid EI+KG. 
We empirically benchmark these methods against existing methods and demonstrate their superiority over the state-of-the-art.
\end{abstract}



%\received{June 2024}
%\received[revised]{June 2024}
%\received[accepted]{June 2024}

\maketitle

\section{Introduction}

Optimising real-world problems often involves evaluating objective functions that are expensive—either financially, temporally or computationally. For instance, determining the optimal shape of an aerofoil requires assessing the performance of a proposed design using computational fluid dynamics (CFD) simulations, which can take several hours to complete~\cite{morita2022applying}. Due to the high cost of each evaluation, performing more than a few hundred simulations to identify an optimal design is rarely feasible.

Bayesian optimisation (BO) has become a widely used strategy for addressing optimisation problems involving expensive objective functions~\cite{shahriari2015taking}. The process typically starts by constructing a surrogate Bayesian model -- most commonly a Gaussian process (GP) -- of the objective function, using an initial dataset generated via low-discrepancy design of experiments methods such as Latin Hypercube sampling~\cite{mckay1979comparison}. The search for optimal solutions is then directed by an acquisition function, which carefully balances exploitation (selecting candidates predicted to perform well) and exploration (targeting regions with high predictive uncertainty)~\cite{de2021greed}. This approach enables the identification of the most promising candidates for costly evaluation. After each new evaluation, the dataset is updated, and the process repeats until the allocated budget for expensive evaluations is exhausted. Bayesian optimisation has consistently demonstrated strong performance in tackling complex, real-world optimisation tasks \cite{morita2022applying}.

A significant portion of research in Bayesian optimisation (BO) has focused on developing effective acquisition functions. For a comprehensive overview, readers are encouraged to consult e.g.~\citet{shahriari2015taking}. Two acquisition strategies particularly relevant to this discussion are expected improvement (EI) and the knowledge gradient (KG). The most widely used acquisition function is expected improvement, originally introduced by \citet{mockus:ei}, and later popularised by \citet{Jones1998EfficientFunctions} as part of an algorithm referred to as Efficient Global Optimization~(EGO). The EI function selects the next candidate for expensive evaluation by identifying the solution that, on average, is expected to yield the greatest improvement over the current best, accounting for predictive uncertainty. 
While EI provides a sensible balance between exploration and exploitation, it does so myopically: it selects the point that maximises the immediate expected improvement given the current GP posterior. In contrast, the Knowledge Gradient (KG) acquisition function~\citep{Scott2011TheRegression}  explicitly accounts for the expected impact of a new observation on the GP model. KG selects the candidate whose evaluation would yield the largest expected increase in the best predicted objective value after updating the GP model with the newly sampled data.

%While EI is considered Pareto compliant with respect to the exploration–exploitation trade-off~\cite{de2021greed}, it does not account for changes in GP hyperparameters that occur when new data points are added to the dataset. The knowledge gradient acquisition function addresses this limitation by aiming to identify the candidate whose evaluation would result in the largest expected improvement in the model’s best mean prediction, after incorporating the new data. 
%and marginalising over possible hyperparameter values. 
%This approach directly accounts for the fact that both the hyperparameters and the predictive distribution are updated as new data points are added.

% in this uncostrained setting, the efifcacy of amny acuisitions functions have been explored; see for example, 

% During the last decade, BO has been extended to the constrained setting. For instance, constrained Expected Improvement (cEI) considers the EI acquisition function weighted by the probability of feasibility \cite{Gardner2014BayesianConstraints}. More recently, \citet{Ungredda2024BayesianProblems} explored constrained Knowledge Gradient (cKG)  which extends KG to the constrained problem. Other methods make use of the Augmented Lagrangian \cite{Picheny2016BayesianLagrangian} which converts the constrained problem into an unconstrained one, or information-theoretic methods such as Predictive Entropy Search with constraints (PESC) \cite{MiguelHernandez-Lobato2015PredictiveConstraints}.

In recent years, there has been growing interest in addressing expensive constraints within Bayesian optimisation (BO); for a comprehensive overview, see the work of \citet{amini2024constrained}. Usually, a separate Gaussian Process surrogate model is built for each constraint, which can then be used to predict the probability of feasibility of a candidate solution A prominent example of an acquisition function for constrained BO is constrained Expected Improvement (cEI), which adapts the standard EI acquisition function by weighting it according to the probability that a candidate solution is feasible \cite{Schonlau1998GlobalModels,Gardner2014BayesianConstraints}. More recently, \citet{Ungredda2024BayesianProblems} proposed the constrained Knowledge Gradient (cKG), which extends the KG framework to directly incorporate constraints.

Most of the literature on constrained Bayesian optimisation assumes that the objective and all the constraint functions are evaluated jointly, e.g., because there is a single experiment or simulation model that results in all relevant values. In practice, however, constraints are often evaluated by a separate model, i.e., it is possible to evaluate only a subset of the constraints, and different functions may require varying amounts of time to evaluate. Even when runtimes are similar, some functions may be easier to learn and require fewer evaluations than others. Allowing a decoupling of the evaluations -- allowing objective and constraint functions to be assessed independently -- enables the optimisation process to focus computational resources on the most informative evaluations. This leads to more data- and computation-efficient learning and optimisation, particularly when some functions are significantly more expensive or complex than others.

% A potential pitfall of these methods is that they evaluate all the black-box functions together (coupled evaluation) and cannot exploit decoupled evaluations. It might be interesting to consider the case where black-box functions are evaluated separately. Decoupled constrained BO was first considered by \citet{Gardner2014BayesianConstraints}, where they describe the ``chicken-and-egg pathology'' of cEI for decoupled evaluations. PESC was also extended to decoupled evaluations quite trivially as its acquisition function can be decomposed into individual functions \cite{MiguelHernandez-Lobato2015PredictiveConstraints, MiguelHernandez-Lobato2016ASearch}. \citet{Nguyen2024OptimisticConstraints} utilise the popular principle of ``optimism in the face of uncertainty'' from multi-armed bandits \cite{Papini2019OptimisticSampling} and propose an algorithm with a provable performance guarantee in the decoupled setting.

\citet{Gelbart2014BayesianConstraints} were the first to identify the challenges associated with decoupled evaluations in constrained Bayesian optimisation. Expanding on this, \citet{Hernandez-Lobato2015PredictiveConstraints} proposed the first principled solution: Predictive Entropy Search for Constrained problems (PESC). This method iteratively selects candidate solutions and evaluates either a constraint or the objective function -- rather than both simultaneously -- to reduce uncertainty around the optimum considering the feasibility constraints.

% A more recent contribution by \citet{Nguyen2024OptimisticConstraints} introduced UCB-D, a variant of the Upper Confidence Bound (UCB) algorithm adapted for constrained settings. In the unconstrained context, UCB combines the mean prediction of a candidate with its predictive uncertainty to guide selection. In UCB-D, the feasible space is relaxed, and the objective function is evaluated only if the candidate lies within this relaxed region. Otherwise, the algorithm focuses on evaluating the most violated constraint. This approach has shown strong empirical performance and is supported by theoretical guarantees.

A more recent contribution by \citet{Nguyen2024OptimisticConstraints} introduced UCB-D, a variant of the Upper Confidence Bound (UCB) algorithm adapted for constrained settings. In the unconstrained context, UCB combines the mean prediction of a candidate with its predictive uncertainty to guide selection. In UCB-D, after selecting a candidate location by maximizing the objective upper confidence bound over an optimistic feasible region, the algorithm compares the maximum constraint violation risk against the objective’s uncertainty term. If the constraint violation risk dominates, UCB-D queries the most violated constraint; otherwise, it queries the objective function.

Despite these developments, the Knowledge Gradient (KG) framework has not yet been explored for decoupled evaluations in constrained optimisation. In this paper, we address this gap by introducing new methods based on cKG, and cEI, and by experimentally evaluating their effectiveness against leading decoupled approaches -- PESC -- from the literature.

% the two leading decoupled approaches -- PESC and UCB-D -- from the literature.

% In this report, we propose three new constrained BO methods for decoupled evaluations: decoupled constrained Knowledge Gradient (dcKG), decoupled constrained Expected Improvement (dcEI) and a hybrid method between cEI and dcKG (EI+KG) and explore their performance on two different synthetic problems.

The remainder of this paper is organised as follows. Section \ref{sec:prob} introduces the problem setting and provides a brief overview of constrained Bayesian optimisation. Section \ref{sec:methodology} presents our original contributions, where we extend existing acquisition strategies to the decoupled constrained setting. In Section~\ref{sec:results}, we report numerical experiments that evaluate the performance of the proposed methods on synthetic benchmark problems. Finally, Section \ref{sec:con} concludes the paper and outlines key open questions for future research.

% \section{Problem Setting and Bayesian Optimisation}
\section{Background}
\label{sec:prob}
\subsection{Problem Setting}

% We consider a sub-region $X \subset \mathbb{R}^{d},$ and a black-box function $f : X \rightarrow \mathbb{R}$  with constraints $c_k:X \rightarrow \mathbb{R}, k=1,\dots, K$. We wish to find the optimal design vector, $x^*$, of this function subject to each constraint function being negative:

Let $x \in \mathcal{X} \subset \mathbb{R}^d$ be a $d$-dimensional design vector within a feasible space $\mathcal{X}$, defined by a set of $K$ constraint functions $c_k: x \rightarrow \mathbb{R}$, for $k \in \{1, \dots, K\}$. The objective is to maximise a function $f: x \rightarrow \mathbb{R}$ over the feasible region. Formally, we seek the optimal solution:

\begin{equation}
     {
     x^*=\underset{x \in \mathcal{X}}{\arg \max} \; f(x), \quad \text{s.t.}  \quad c_k(x) \leq 0  \quad \forall \; k=1, \dots, K.}
\end{equation}

% The objective function $f$ takes as an input a design vector $x \in X$ and returns an observation (which we assume to be noiseless) $y = f (x)$ or $y=c_{k}(x), k=1,\dots, K$, where we assume that constraint and objective function observations are independent and may be evaluated individually. 

% The latter assumption contrasts with the more common assumption in the literature that if we evaluate a design $x,$ then we must also evaluate \textit{all} of the constraints at that point, and hence receive the objective function value and a \textit{vector} of constraint values. \citet{MiguelHernandez-Lobato2015PredictiveConstraints} gives the example of a financial simulator, in which every part of a simulation must be run at once.

% We assume that we have a total budget of $B$ samples, and after the budget has been consumed we should return a recommended design, $\bx_r$. Its quality is determined by the difference between the true optimal value, $f(\bx^*),$ and the value of the objective function at our recommended point, $f(\bx_r).$ We naturally come to the question of how to penalise infeasibility, because we are only interested in those points which are feasible, i.e. $x \in  F$, where $F = \{x|c_k(x) \leq  0; \forall k = 1,\dots, K\}$. If $x_r$ is not feasible, then we simply set the objective function to return some penalty value, $M$. Finally, we measure the quality of a solution by an Opportunity Cost (OC) to be minimised:

We consider a decoupled problem setting where the objective and each constraint are expensive and may be evaluated independently, and cost of evaluating each of them may differ. We denote the cost of evaluating the objective as $b_0$ and the cost of evaluating constraint $k$ as $b_k$; where $k=1,2, \ldots K$. The cost of one \emph{coupled} evaluation is therefore $\sum^{K}_{k=0}{b_k}$. The total computational expense incurred during optimization can be computed by simply multiplying the number of objective/constraint evaluations by their respective costs and adding them. Note that for expensive optimization problems, the computation cost incurred in other algorithmic operations~(e.g. infill search) is considered negligible.

We assume a total budget of $B$. Once this budget is exhausted, the algorithm recommends a design vector $x_r$. The quality of this recommendation is assessed by comparing it to the true optimum $x^*$.

If $x_r$ is feasible, the quality is measured as the difference in objective value between $x^*$ and $x_r$. If $x_r$ is infeasible, the quality is instead the difference between the objective value at $x^*$ and a penalty value M. This quality measure is referred to as the \emph{opportunity cost (OC)}, and is formally defined as:

\begin{equation}
    \centering
        \text{OC}(x_r) =
        \begin{cases}
            f(x^*) - f(x_r),& x_r \in F.\\
            f(x^*) - M,    & x_r \notin F. \\
        \end{cases}
    \label{eq:OC}
\end{equation}

% Without loss of generality, we assume that the penalty $M$ is equal to zero. However, $M$ may be set by using the minimum surrogate model estimate of the objective function in the design space \cite{Letham2019ConstrainedExperiments}. We also remark that the penalty value is often best set by an application expert \cite{Ungredda2024BayesianProblems}. 

Without loss of generality and for notational simplicity, unless specified otherwise, we assume that the penalty $M$ is equal to zero. However, $M$ may also be set using the minimum surrogate model estimate of the objective function over the design space~\cite{Letham2019ConstrainedExperiments}. In our experiments, we adopt this approach and explicitly state the value of $M$ used in each case. We also note that the choice of penalty is often best made by an application expert, as discussed by~\citet{Ungredda2024BayesianProblems}.

% \subsection{Summary of Bayesian Optimisation}

% As stated earlier, we build a surrogate model and then use this model to quantify the added value given by evaluating each candidate point in the task of determining the location of the optimiser. We introduce here the notation and basic concepts that we use in our report. The basic procedure for BO is as follows:\\

% \begin{center}
% \begin{algorithm}[H]
% Perform some space-filling initial data collection\\
% \While{budget B of evaluations is not exhausted}
% {
% Fit independent GPs for each constraint and the objective\\
% Use the acquisition function to obtain the most promising location (and constraint/objective in the decoupled case) to evaluate next\\
% Evaluate the true quality of this design\\
% Update the surrogate models with the new evaluated point\\
% }
% Return the recommended design

% \caption{Bayesian optimization procedure}
% \end{algorithm}
% \end{center}

\subsection{Gaussian Processes}

BO relies on surrogate models to approximate the relationships between design vectors and the responses of both constraint and objective functions—modelled independently. The most commonly used surrogate is the \emph{Gaussian Process (GP)} model, which offers two key advantages:
\begin{itemize}
    \item It provides a \textbf{mean prediction} that serves as a best guess of the function value, useful for exploitation.
    \item It quantifies \textbf{predictive uncertainty}, which guides exploration.
\end{itemize}

Additionally, the GP’s predictive distribution is Gaussian, enabling analytical computation of expected utility for evaluating candidate solutions. This often makes it computationally tractable to identify promising designs efficiently.

Typically, we begin with a set of $n$ initial design points, $X = (x_i)_{i=1}^n \in \mathbb{R}^{(n\times d)}$, selected using a space-filling method such as \emph{Latin Hypercube sampling}~\cite{mckay1979comparison}. Each design is evaluated across all constraint and objective functions, yielding a dataset $D = \left(x_i, c_1(x_i), \dots, c_K(x_i), f(x_i)\right)_{i=1}^n$. Once trained, the GP model for a function produces a predictive distribution:
\[
p(f(x) ~|~ D) \sim \mathcal{N}(\mu(x), \sigma(x)),
\]
where the mean and variance are given by:
\[
\mu_f(x) = \bm{\kappa}(x, X)^\top K_{XX}^{-1} Y,
\]
\[
\sigma_f(x) = \kappa(x, x) - \bm{\kappa}(x, X)^\top K_{XX}^{-1} \bm{\kappa}(x, X).
\]

In the above formulation, $Y \in \mathbb{R}^n$ represents the vector of true function outputs. The covariance matrix $K_{X,X} \in \mathbb{R}^{n \times n}$ is constructed using a kernel function $\kappa(x', x'' \mid \theta)$, which measures the similarity between all pairs of evaluated design points $x', x'' \in X$, conditioned on a hyperparameter vector $\theta$. These hyperparameters are tuned to fit the GP model to the dataset $D$. The vector $\bm{\kappa}(x, X) \in \mathbb{R}^n$ captures the covariance between a new candidate point $x$ and each of the previously evaluated designs in $X$.

The choice of kernel function reflects assumptions about the smoothness and structure of the underlying function being modelled. Common choices include the \emph{Radial Basis Function (RBF)} and the \emph{Matérn 5/2} kernel \cite{Rasmussen2005GaussianLearning}. In this paper, we adopt the Matérn 5/2 kernel and use maximum likelihood for hyperparameter optimisation.

\subsection{Expected improvement acquisition function}
The acquisition function in BO is used to predict how informative a new evaluation would be for finding the optimum. Expected improvement (EI) selects the next evaluation point by maximising the expected improvement over the current best observed objective value. Let $f_{\max}$ denote the best (i.e., maximal) observed function value so far, and let the Gaussian process posterior at a candidate point $x$ have predictive mean $\mu(x)$ and standard deviation $\sigma(x)$. The improvement at $x$ is defined as $I(x) = \max\{ f_{\max} - f(x), 0 \}$. Since $f(x)$ is modelled by the GP as normally distributed, the expected improvement has the closed-form expression
\begin{eqnarray}
\mathrm{EI}(x) = \mathbb{E}[I(x)]
= (\mu(x) - f_{\max}) \, \Phi\!\left( \frac{\mu(x) - f_{\max}}{\sigma(x)} \right)
	+	\sigma(x) \, \phi\!\left( \frac{\mu(x) - f_{\max}}{\sigma(x)} \right),
    \end{eqnarray}
where $\Phi(\cdot)$ and $\phi(\cdot)$ denote the standard normal cumulative distribution function and probability density function, respectively. This formulation naturally balances exploitation (sampling near the current best) and exploration (sampling where uncertainty is high).

\subsection{Predicting feasibility and recommending a solution}

When the predictive distribution of each constraint function $p(c_k(\bx) \mid D)$ is Gaussian, the \emph{probability of feasibility} for a candidate design $\bx$ with respect to constraint $k$ is given by:
\[
\text{PF}_k(x) = \mathbb{P}\left(p(c_k(x) \mid D) \leq 0\right) = \Phi\left(-\frac{\mu_{c_k}(x)}{\sigma_{c_k}(x)}\right),
\]
where $\Phi(\cdot)$ denotes the cumulative distribution function (CDF) of the standard normal distribution.

Since each constraint is modelled independently, the overall probability that $x$ is feasible—i.e., satisfies all constraints—is simply the product of individual feasibility probabilities:
\begin{equation}
\text{PF}(x) = \prod_{k=1}^K \text{PF}_k(x).    
\end{equation}

% This fact gives the natural generalisation of $x_{r}$ to the constrained setting as $$x_{r} = \arg\max_{x\in X} \mu_y^B(x)\mathbb{P}[ \mathbf{c}(x) \leq 0].$$

This fact gives the natural generalisation of $x_{r}$ to the constrained setting as:

\begin{equation}
    x_{r} = \arg\max_{x\in X} \mu_f(x) \text{PF}(x).
\end{equation}

\section{Methodology \label{sec:methodology}}
In this section, we introduce two new BO algorithms for decoupled constrained problems. The first one is a fully decoupled version of the constrained knowledge gradient (cKG) approach proposed in \cite{Ungredda2024BayesianProblems}. The second is an extension to constrained EI which uses cEI to determine the next solution to be evaluated, but the decoupled constrained KG idea  to decide which function to evaluate. The latter approach is theoretically not as principled, but computationally faster. Because both ideas are based on the cKG approach, we start first in Section~\ref{sec:cKG} by providing a summary of cKG, before explaining our new decoupled approaches in Sections~\ref{sec:dcKG} and \ref{sec:cEIplus}.

\subsection{Constrained Knowledge Gradient\label{sec:cKG}}
The idea of the \textit{unconstrained} Knowledge Gradient (KG) acquisition function \cite{Scott2011TheRegression} is to compute the expectation of the difference between the  maximum of the current posterior mean of the GP surrogate model, $\mu^n(.)$, and the maximum of the new posterior mean, $\mu^{n+1}(.)$, after we sampled at a point $x$, i.e.,

\begin{equation}
\label{KG}
\text{KG}(x)=\mathbb{E}\left[ \max_{x' \in X} \{\mu_y^{n+1}(x') \} - \max_{x' \in X} \{ \mu_y^n(x') \}  \vert x_{n+1}=x\right].
\end{equation}

As this cannot be computed analytically, different numerical approaches have been proposed, see \citet{Ungredda2022EfficientOptimization}.

In \cite{Ungredda2024BayesianProblems}, KG was extended to the case of (fully coupled) constraints by learning separate GPs for each constraint, and considering the impact of an evaluation on the quality of the recommended solution, where the recommended solution is the solution with the best posterior mean, multiplied by the probability that it is feasible:
\[
x^n_r = \arg\max_{x \in X} \mu^n_y (x) \text{PF}^n(x).
\]

The difference in performance between the current recommended design and the new best performance results in the following acquisition function:
\begin{equation}
\text{cKG}(x)=\mathbb{E}\left[ \max_{x' \in X} \{ \mu_y^{n+1}(x')\text{PF}^{n+1}(x')\} - \mu_y^{n+1}(x_r^n)\text{PF}^{n+1}(x_r^n) \vert x_{n+1}=x \right].
\end{equation}

By Jensen's inequality, the last equation is positive for all the design space and \citet{Ungredda2024BayesianProblems} pointed out that $\mu_y^{n+1}(x_r^n)$ may be marginalised over $y^{n+1}$, such that
\begin{equation}
\text{cKG}(x)=\mathbb{E}\left[ \max_{x' \in X} \{ \mu_y^{n+1}(x')\text{PF}^{n+1}(x')\} - \mu_y^{n}(x_r^n)\text{PF}^{n+1}(x_r^n) \vert x_{n+1}=x \right] .
\end{equation}

This acquisition function quantifies the benefit of the next point at which we sample taking into account not only the mean weighted by probability of feasibility, but also how new information about the constraints may affect our estimation of the probability of feasibility of our existing $x_{r}$.
Note that if there are no constraints, it reduces to the standard KG.

\subsection{Decoupled Constrained Knowledge Gradient (dcKG)\label{sec:dcKG}}

Now let us introduce the modification to cKG that will allow us to handle decoupled constraints. 
In order to present decoupled constrained KG (dcKG), let us introduce a variable $m$ which counts the iteration number, rather than function evaluations for a given constraint, which may differ ($m$ effectively counts total function evaluations). We then define $q(m)$ to be the index of the constraint or the objective function which we evaluate at a given iteration $m$. Our acquisition function is then, \begin{equation}
\text{dcKG}^k(x)=\mathbb{E} \left[ \max_{x' \in X} \left\{ \mu_y^{m+1}(x')\text{PF}^{m+1}(x')\right\}  - \mu_y^{m+1}(x_r)\text{PF}^{m+1}(x_{r}^{n}) \big\vert x_{m+1}=x, q(m+1) = k \right] \times \frac{1}{b_k},
\end{equation}
where $k \in \{0, ..., K \}$ and we treat $\text{dcKG}^0(x)$ to correspond to the KG value obtained from sampling the objective function.

However, considering a set of $K+1$ optimizers that only evaluate the benefit of sampling a single source without taking other functions into account brings an issue, discussed in \citet{Gelbart2014BayesianConstraints}. It is referred as the "chicken and egg" pathology in the context of expected improvement where the decoupled acquisition function stops sampling unexplored regions and gives preference to known and feasible regions of $X$. 

For dcKG, consider the case of a single objective and constraint with a well-explored feasible region. However, the global optimum is located in a region  of $X$ yet to be discovered. If we attempt to optimise the decision of sampling only the objective, then we would not improve our current model for $\text{PF}(x)$ and the potential improvement in objective value may not exceed the current best estimation in the well-explored region. Conversely, if we tried to only sample the constraint, we would not see the benefit of possibly finding better observations for the objective. As a result, the objective function estimate would remain close to the prior, causing the acquisition function to concentrate once again on the well-known region. 

To address this issue, we include cKG as an additional aquisition function to evaluate the joint improvement of selecting \emph{both} the objective \emph{and} all of the constraints, divided by the total cost of evaluating them all. If the largest value is found by cKG, in order to still benefit from the possibility of decoupling, we will at the identified location $x$
not evaluate constraint functions that have a high probability of being feasible at that location, given a threshold $\text{PF}(x) < 1 - \delta$. In all experiments, we assume a fixed threshold value $\delta = 1.0 \times 10^{-7}$. The process of one  iteration is summarised in the pseudocode below.

\begin{algorithm}[htb]
\caption{Decoupled Constrained Knowledge Gradient (dcKG)}
% Fit Gaussian Process models using current data
Fit GP models based on the current dataset $D$\;
Initialize an empty list \texttt{DKG\_values} = [ ]\;

\For{$k \in (0,\,K+1)$}{
    % Evaluate decoupled KG for each source k using L-BFGS
    $(x^*_k, \text{dckg}^*_k) = \text{Optimise}(\text{dcKG}^k)$  \tcp*[r]{Optimized with L-BFGS}
    Append $\text{dckg}^*_k$ to \texttt{DKG\_values}\;
}

% Evaluate coupled KG candidate
Compute candidate $(x^*, \mathrm{cKG}^*)$ by optimizing the coupled KG acquisition function\;

\If{$\mathrm{cKG}^* > \max(\texttt{DKG\_values})$}{
    % Identify constraint functions that are not likely feasible
    Evaluate the constraint functions for all $k \in  \{\, k \in \{1,\dots,K\} \mid \mathrm{PF}^k(x^*) < 1 - \delta \,\}$\;
    Evaluate the objective function at $x^*$\;
    Update dataset $D$\;
}
\Else{
    % Select the candidate with the highest dKG value if coupled KG is lower
    Identify the source $k'$ corresponding to $\max(\texttt{DKG\_values})$\;
    Evaluate the corresponding constraint or objective function at $x^*_{k'}$\;
    Update dataset $D$\;
}
Return the updated dataset $D$ with the new evaluations\;
\end{algorithm}

\subsection{Constrained Expected Improvement plus Decoupled Evaluation (cEI+) \label{sec:cEIplus}}
In order to extend cEI to the decoupled case, we propose to use cEI to determine the \emph{location} $x$ to evaluate, but then use dcKG($x$) to determine which constraint or objective function to evaluate at this location. We denote this approach cEI+. 

One would not expect this method to outperform the full dcKG algorithm, because the point finally chosen to evaluate is not optimised for the constraint we end up evaluating. However, it is much less computationally expensive than the dcKG because at each iteration we only need to calculate the dcKG values at a single point, rather than at many points in order to optimise it. 

\section{Results and discussion}
\label{sec:results}

In this section, we conduct numerical experiments to quantify the performance of the two decoupled approaches proposed, and benchmark them with some of the existing methods in BO. We start with baseline comparisons assuming equal costs for each objective/constraint. Then we extend the results to modified cases, including redundant constraints and heterogeneous costs~(where objective and constraints may have different costs). Lastly, we provide an ablation study on dcKG, the main algorithm proposed in this work. 

\subsection{Baseline experiments}

For these experiments, we consider a few well-known test functions used in the field of BO, including Branin, Mystery and Test function 2 \cite{Sasena2002FlexibilityApproximations, Ungredda2024BayesianProblems}. Branin and Mystery functions have a single non-linear constraint each which is active at the optimum. Test function 2 has three constraints, of which two are active at the optimum. The formulations of these problems are included in the appendix. For benchmarking, we consider the following approaches from the literature:

\begin{itemize}
    \item Constrained Knowledge Gradient (cKG)~\cite{Ungredda2024BayesianProblems} 
    \item Constrained Expected Improvement (cEI)~\cite{Schonlau1998GlobalModels,Gardner2014BayesianConstraints} 
    \item Predictive Entropy Search with Constraints~(PESC)~\cite{Hernandez-Lobato2015PredictiveConstraints}
%   \item Upper Confidence Bound Gaussian Process (UCB-D)~\cite{Nguyen2024OptimisticConstraints}
\end{itemize} 
  
Among these, cKG and cEI handle the constraints in a coupled manner, while PESC is able to handle decoupled constraints. cKG was implemented using code provided in the GitHub of \cite{Ungredda2024BayesianProblems}, and cEI was implemented using BoTorch \cite{Balandat2020BoTorch:Optimization}. We use the original implementation of PESC in Spearmint \cite{Hernandez-Lobato2015PredictiveConstraints}. %UCB-D was reimplemented using BoTorch.

The computational cost of each constraint and objective function are considered to be equal~(1 unit each). We use 6 initial design samples for which all constraints and the objective are evaluated. The computational budget is set equivalent to 150 coupled evaluations. The results are based on 50 replications with different random seeds.

\begin{figure}[!ht]
    \centering
    \includegraphics[width=\linewidth]{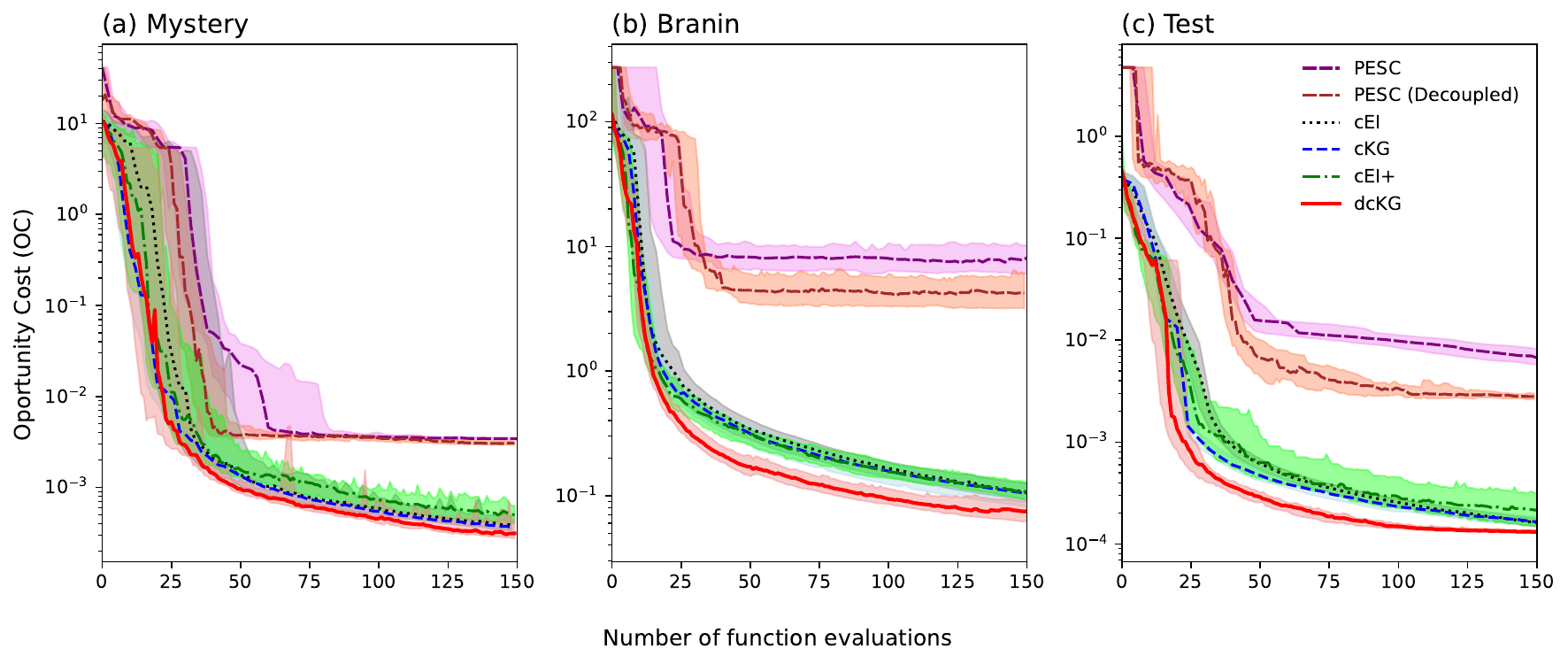}
    \caption{Opportunity cost of the different acquisition functions in terms of the number of function evaluations. \textbf{(a)} Mystery. \textbf{(b)} Constrained Branin. \textbf{(c)} Test Function 2.} 
    \label{fig:synthetic}
\end{figure}

The median value of the opportunity cost~(OC) across the runs, along with 25 and 75 percentile values, are shown in Fig.~\ref{fig:synthetic}. It can be seen that the proposed dcKG converges fastest in all three cases. PESC, though designed for handling decoupled constraints, exhibit slowest convergence among the compared methods, and notably also worse than the standard coupled constraint approaches cEI and cKG. The cEI+ method shows mixed performance, being marginally better than coupled approaches for Branin and marginally worse for the other two. 

\begin{figure}[!ht]
    \centering
    \includegraphics[width=\linewidth]{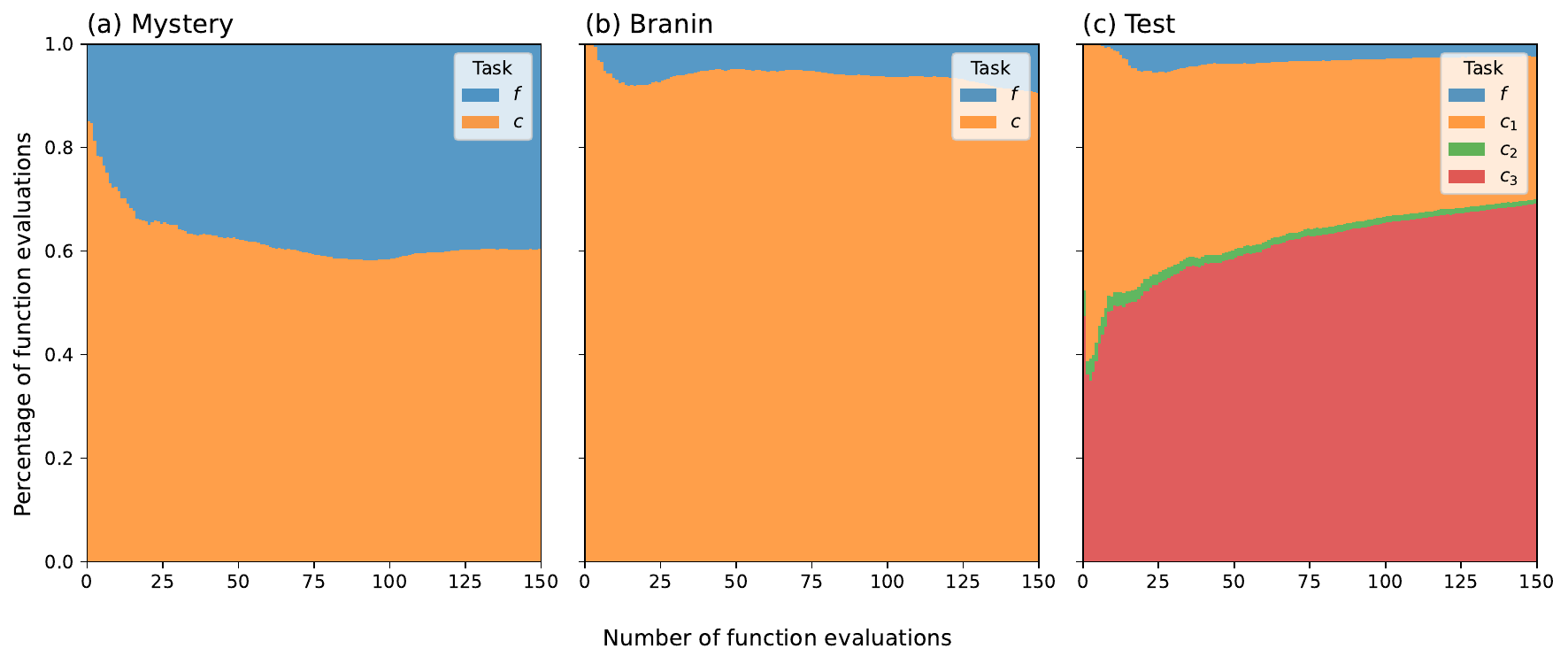}
    \caption{Cumulative evaluations for each different function (task), using dcKG. \textbf{(a)} Mystery. \textbf{(b)} Constrained Branin. \textbf{(c)} Test Function.}
    \label{fig:synthetic_dckg_evals}
\end{figure}

\begin{figure}[!ht]
    \centering
    \includegraphics[width=\linewidth]{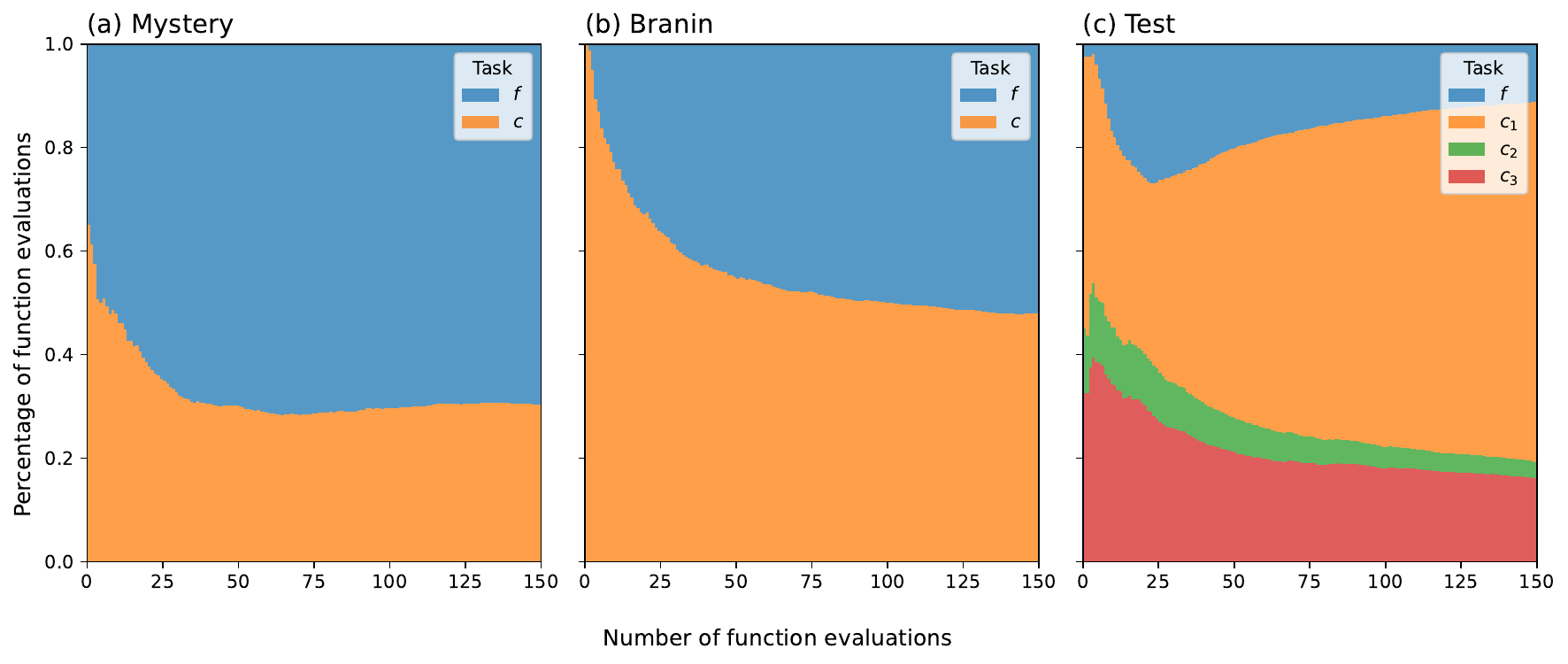}
    \caption{Cumulative evaluations for each different function (task), using cEI+. \textbf{(a)} Mystery. \textbf{(b)} Constrained Branin. \textbf{(c)} Test Function.}
    \label{fig:synthetic_cei_plus_evals}
\end{figure}

In Figs.~\ref{fig:synthetic_dckg_evals} and \ref{fig:synthetic_cei_plus_evals}, we show the cumulative evaluations of each of the functions~(objective and constraint/s) involved in the test problems, for dcKG and cEI+, respectively. Both methods allocate large proportions of computational budget to the constraint and the objective for Mystery function, with dcKG evaluating the constraint more often than cEI+. This is expected as the feasible area is relatively large, and the method tries to sample near the constraint boundary. For Branin, dcKG allocates significantly higher budget to the constraint evaluation, as the infeasible regions are disconnected and narrow, and higher sampling may be required to learn the landscape. In doing so, dcKG is able to obtain better results than cEI+, which allocates a higher proportion to feasible regions. For  Test Function 2, the Constraints 1 and 3 are active at the optimum, while Constraint 2 is not. Hence, it is expected that these decoupled methods would learn this and choose to sample constraint 2 much less than the other functions. This is verified for both methods in Figs.~\ref{fig:synthetic_dckg_evals}-\ref{fig:synthetic_cei_plus_evals}. It is also seen that similar to the case of Branin, dcKG tends to sample the constraints more often than cEI+, in the process achieving a solution with better objective value at the intersection of the two constraints. 

\subsection{Redundant constraints}

\begin{figure}
    \centering
    \includegraphics[width=0.5\linewidth]{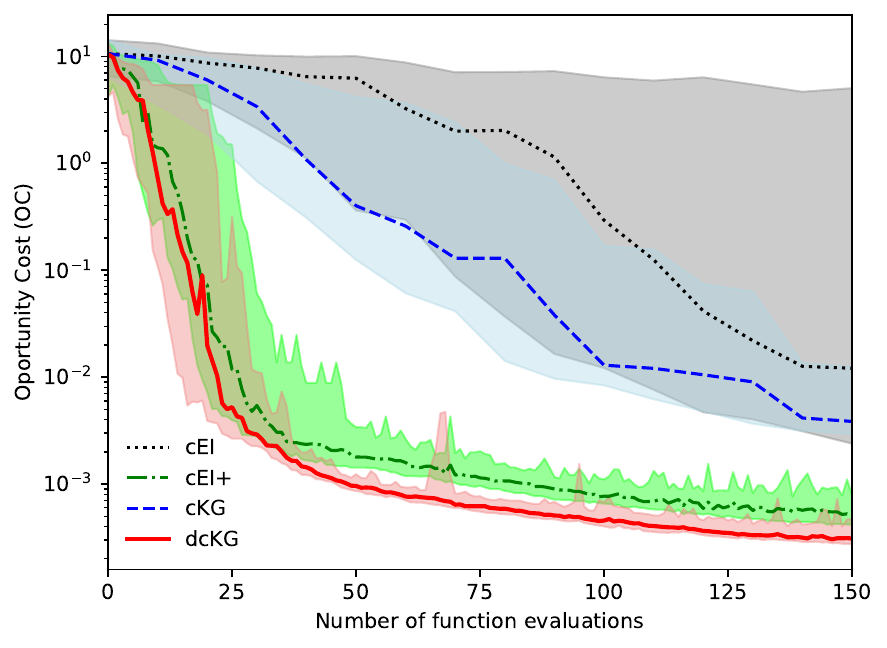}
    \caption{Opportunity cost of the different acquisition functions in terms of the number of function evaluations for the Mystery function with eight dummy constraints.}
    \label{fig:mystery_redundant}
\end{figure}

\begin{figure}
    \centering
    \includegraphics[width=0.7\linewidth]{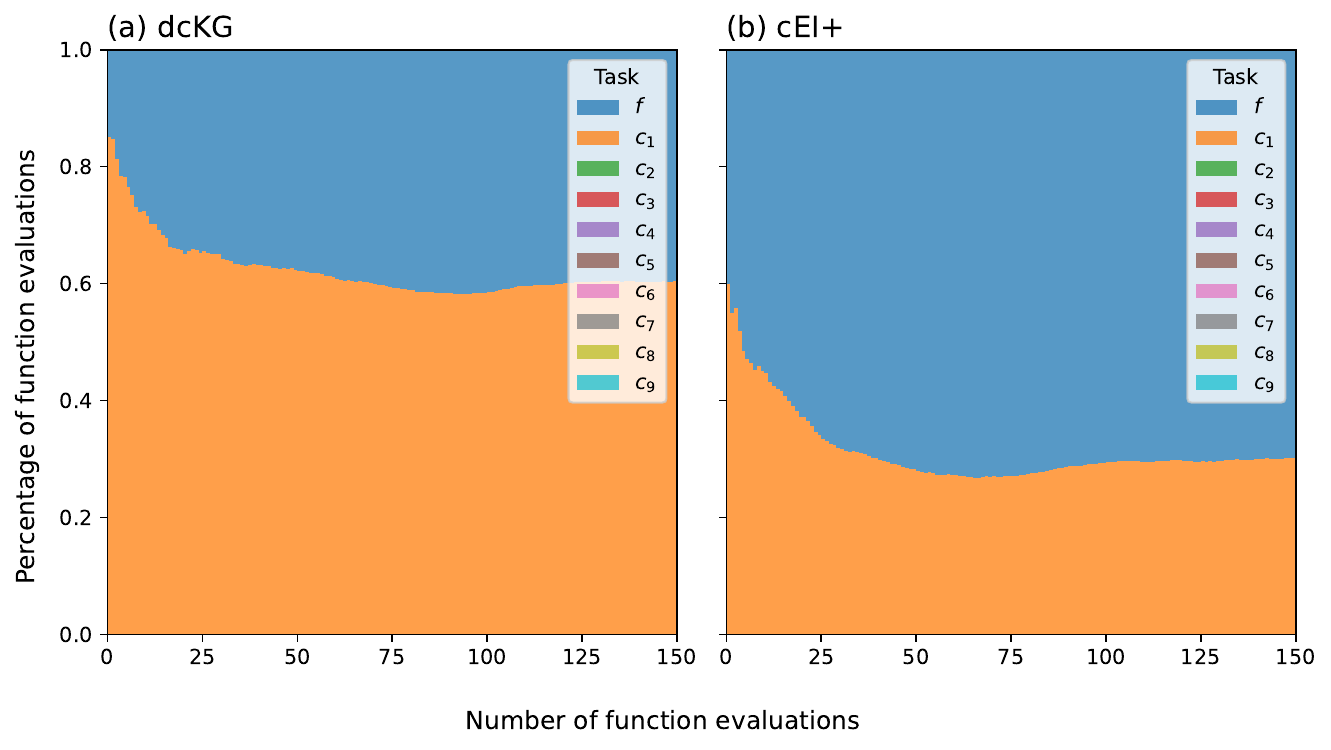}
    \caption{Mystery function with eight dummy constraints: plot of the cumulative evaluations for each different function (task), the initial design is not shown. \textbf{(a)} Using dcKG. \textbf{(b)} Using cEI+.}
    \label{fig:mystery_red_evals}
\end{figure}

In order to further test the capability of the proposed decoupled approaches to sample the relevant constraints, we create a modified version of the Mystery function. We incorporate eight redundant constraints~($c_2-c_9$) in the problem in addition to the original~(active) constraint~($c_1$). These eight constraints are set to a constant value such that they are always satisfied. Fig.~\ref{fig:mystery_redundant} shows the median convergence plot of the OC for the decoupled methods, against the two coupled approaches~(cEI and cKG). Here, it is evident that the decoupled approaches are able to converge much faster through selective evaluation of the relevant constraint instead of sampling the redundant ones. The proportion of evaluations allocated to different constraints and the objective are shown in Fig.~\ref{fig:mystery_red_evals}. It can be seen that after the initial sampling~(not shown in the plot), none of the redundant constraints were sampled by dcKG and cEI+. Thus, the focus was on improving the modeling of the objective and the active constraint in order to achieve a superior solution at each stage of the search compared to the coupled approaches which sample all constraints and the objective unconditionally.  

\subsection{Heterogeneous costs}

In this section, we observe the performance of different algorithms considering objective and constraint(s) may have different evaluation costs.

\begin{figure}[!ht]
    \centering
    \includegraphics[width=0.7\linewidth]{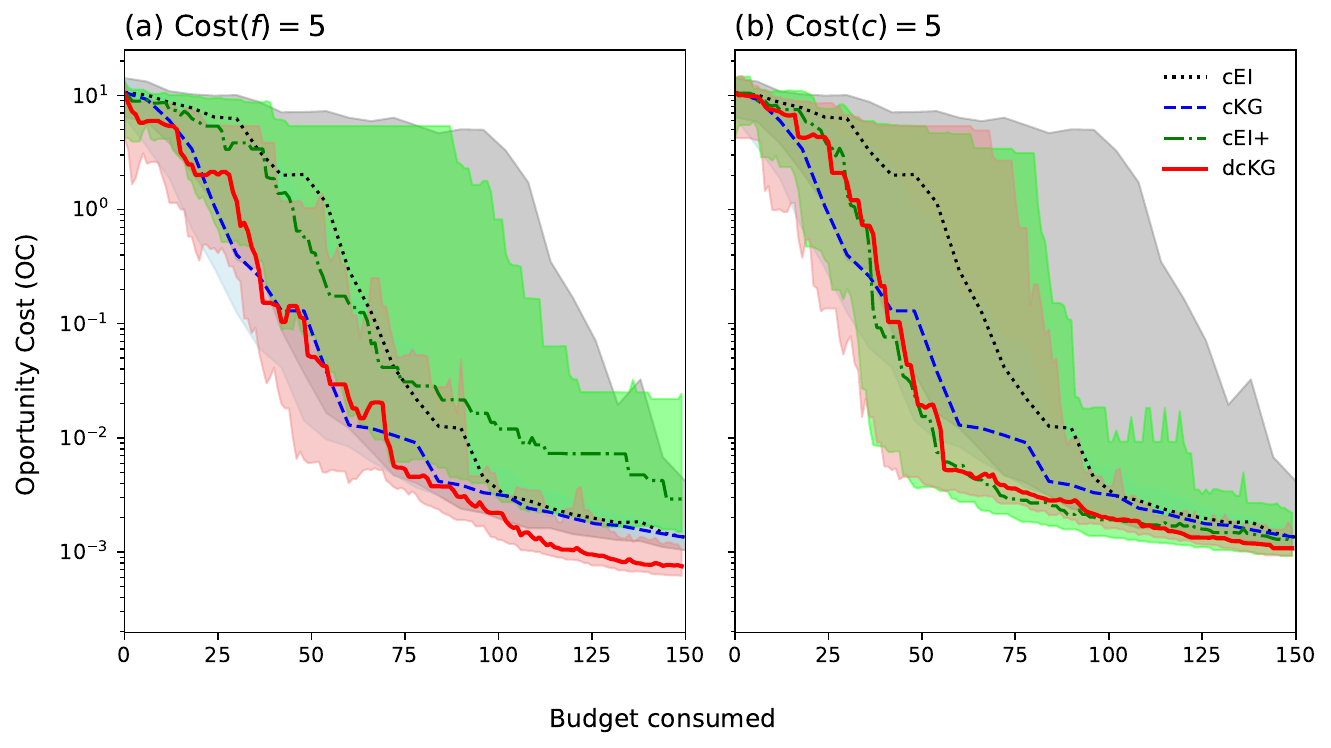}
    \caption{Mystery function: opportunity cost of the different acquisition functions in terms of budget consumed. \textbf{(a)} Cost$(f) = 5$, Cost$(c) = 1$. \textbf{(b)} Cost$(f) = 1$, Cost$(c) = 5$.}
    \label{fig:mystery_costs}
\end{figure}

\begin{figure}[!ht]
    \centering
    \includegraphics[width=0.7\linewidth]{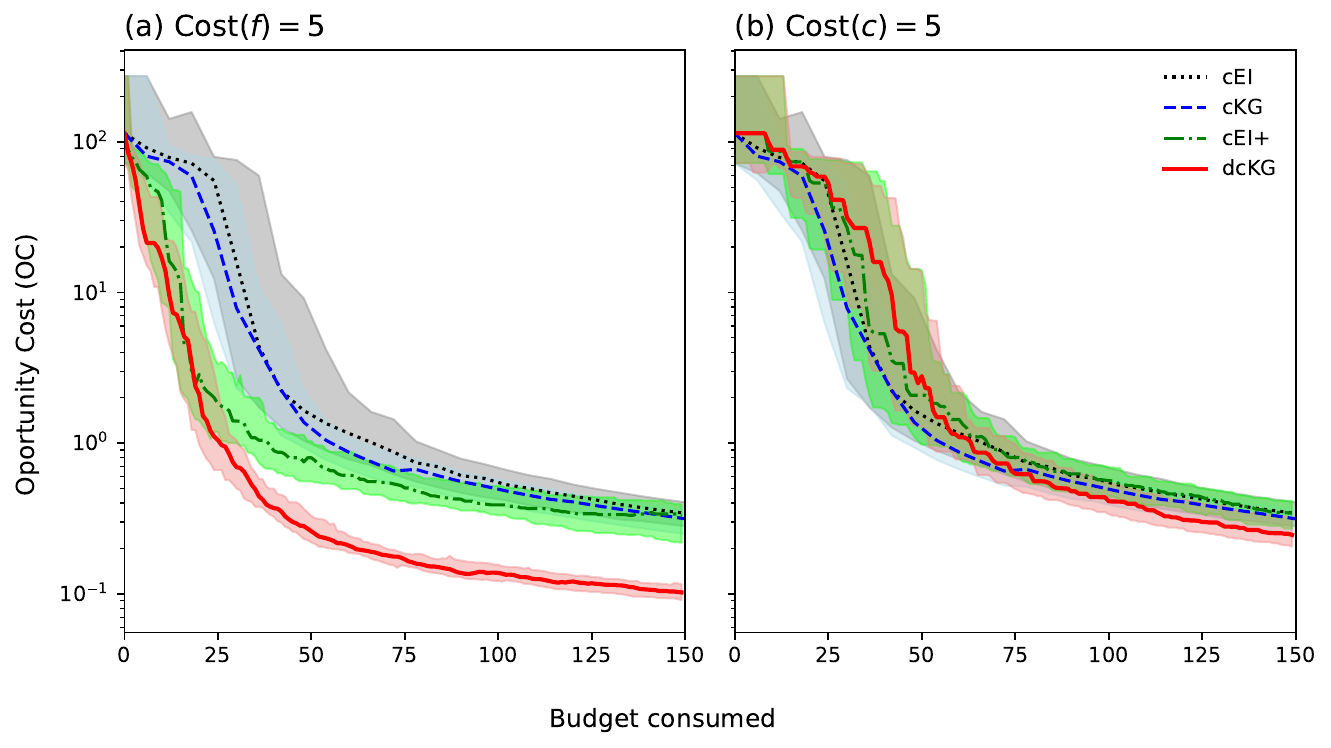}
    \caption{Constrained Branin function: opportunity cost of the different acquisition functions in terms of budget consumed. \textbf{(a)} Cost$(f) = 5$, Cost$(c) = 1$. \textbf{(b)} Cost$(f) = 1$, Cost$(c) = 5$.}
    \label{fig:branin_costs}
\end{figure}

For Mystery and Branin functions, we consider two scenarios - one where evaluation cost of the objective $f$ is five times that of $c$, and vice versa. The results obtained across 50 runs are shown in Fig.~\ref{fig:mystery_costs}~(Mystery) and Fig.~\ref{fig:branin_costs}~(Branin). For  both scenarios in both problems, dcKG exhibits the best overall performance. Notably, the gap in performance relative to other algorithms is larger when $f$ is more expensive to evaluate. This is consistent with the observations in Figs~\ref{fig:synthetic_dckg_evals}-\ref{fig:synthetic_cei_plus_evals}, where dcKG evaluates objectives less often than cEI+. The savings are thus higher when the objective is expensive. Nonetheless, even for the cases where the constraint is expensive, dcKG remains competitive on the objective values. Both coupled algorithms, cEI and cKG show similar performance at the end of the evaluation budget; with cKG generally showing faster convergence at the early stages.   

% \subsection{Suggested experimental results}
% I would suggest to structure the results section as follows:
% \begin{enumerate}
%     \item {\bf Baseline experiments:} Algorithms to compare are dcKG (by which we now mean dcKG+cKG), EI+dcKG, cKG, cEI, PESC. Test functions Branin, Mystery, Test Function 2, a function with several redundant constraints. 
%     \item {\bf Real-world problem:} car side impact problem or one of the other problems mentioned by Hemant. Perhaps we can drop some benchmarks here if too expensive
%     \item {\bf Variable cost: one of the standard benchmarks (perhaps Test Function 2 which has one redundant constraint) but with different cost. We may not need cEI or PESC for this. But it would be good to show that dcKG samples differently depending on the cost (e.g., percentage of evaluations on objective and constraint over the run changes)}
%     \item{Ablation study:} Demonstrate the importance of cKG step, so compare dcKG and dcKG without cKG on a test problem.
% \end{enumerate}

\begin{figure}[!ht]
    \centering
    \includegraphics[width=0.7\linewidth]{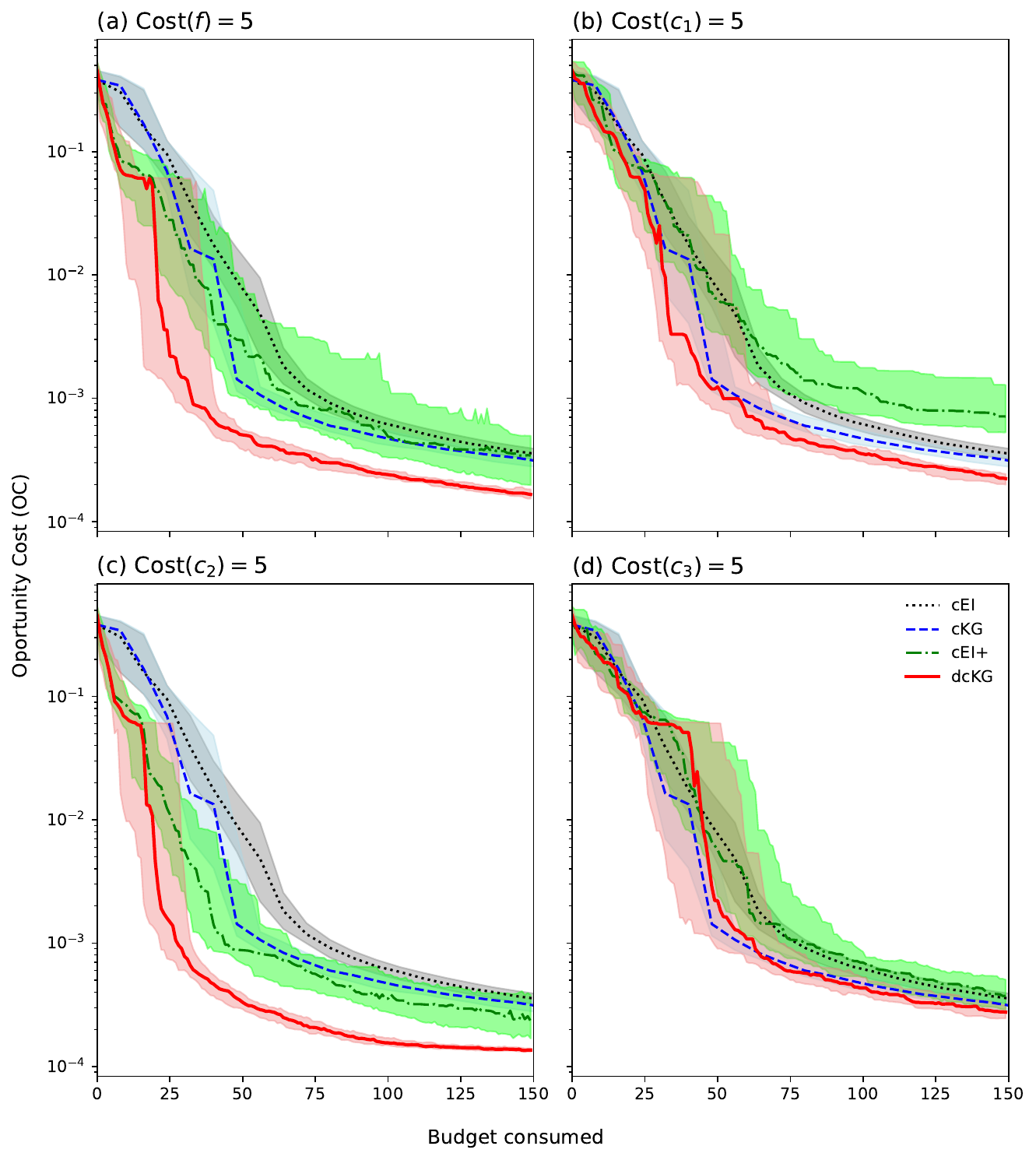}
    \caption{Test function: opportunity cost of the different acquisition functions in terms of budget consumed. \textbf{(a)} Cost$(f) = 5$, Cost$(c_1) = 1$, Cost$(c_2) = 1$, Cost$(c_3) = 1$. \textbf{(b)} Cost$(f) = 1$, Cost$(c_1) = 5$, Cost$(c_2) = 1$, Cost$(c_3) = 1$. \textbf{(c)} Cost$(f) = 1$, Cost$(c_1) = 1$, Cost$(c_2) = 5$, Cost$(c_3) = 1$. \textbf{(d)} Cost$(f) = 1$, Cost$(c_1) = 1$, Cost$(c_2) = 1$, Cost$(c_3) = 5$.}
    \label{fig:test_costs}
\end{figure}

Next, we look at  Test function 2, and consider four scenarios - the objective $f$, or one of the constraints $c_1,c_2,c_3$ incurring cost of 5 units in turn, with others incurring 1 unit each. The results are shown in Fig.~\ref{fig:test_costs}. Once again, dcKG consistently outperforms the other methods scenarios. In particular, its performance is significantly better than others for the case of $\text{Cost}(c_2) = 5$. This is because as evident from Figs.~\ref{fig:synthetic_dckg_evals}-\ref{fig:synthetic_cei_plus_evals}, after learning that $c_2$ is the redundant constraint, dcKG evaluates it much fewer times, including when compared with decoupled approach cEI+.  The coupled approaches perform the worst for this scenario as they evaluate the constraint regardless for all samples. The performance of dcKG is also significantly better than other approaches for the case of $\text{Cost}(f) = 5$, given it spends less computation on  evaluating the objective compared to the (active) constraints. For the remaining two scenarios, where an active constraint~($c_1$ or $c_2$) is expensive, dcKG still maintains its competence in achieving lower opportunity cost. Evidently, spending more evaluations on the constraints that matter~(are active) helps dcKG to learn them more accurately, in turn yielding a more precise location of their intersection where the true optimum is located. 

\subsection{Ablation study}

\begin{figure}
    \centering
    \includegraphics[width=0.7\linewidth]{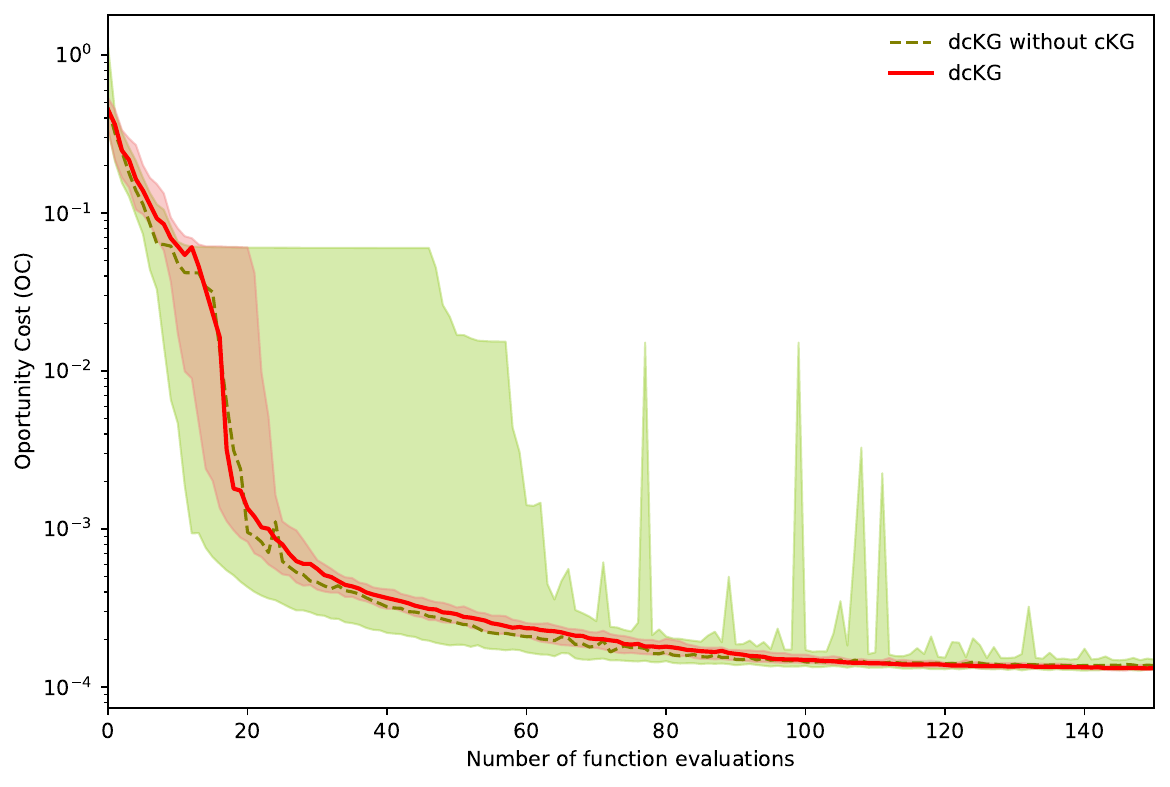}
    \caption{Test function: opportunity cost in terms of the number of function evaluations, showing dcKG (red) and dcKG without cKG (green).}
    \label{fig:ablation}
\end{figure}

\begin{figure}
    \centering
    \includegraphics[width=0.7\linewidth]{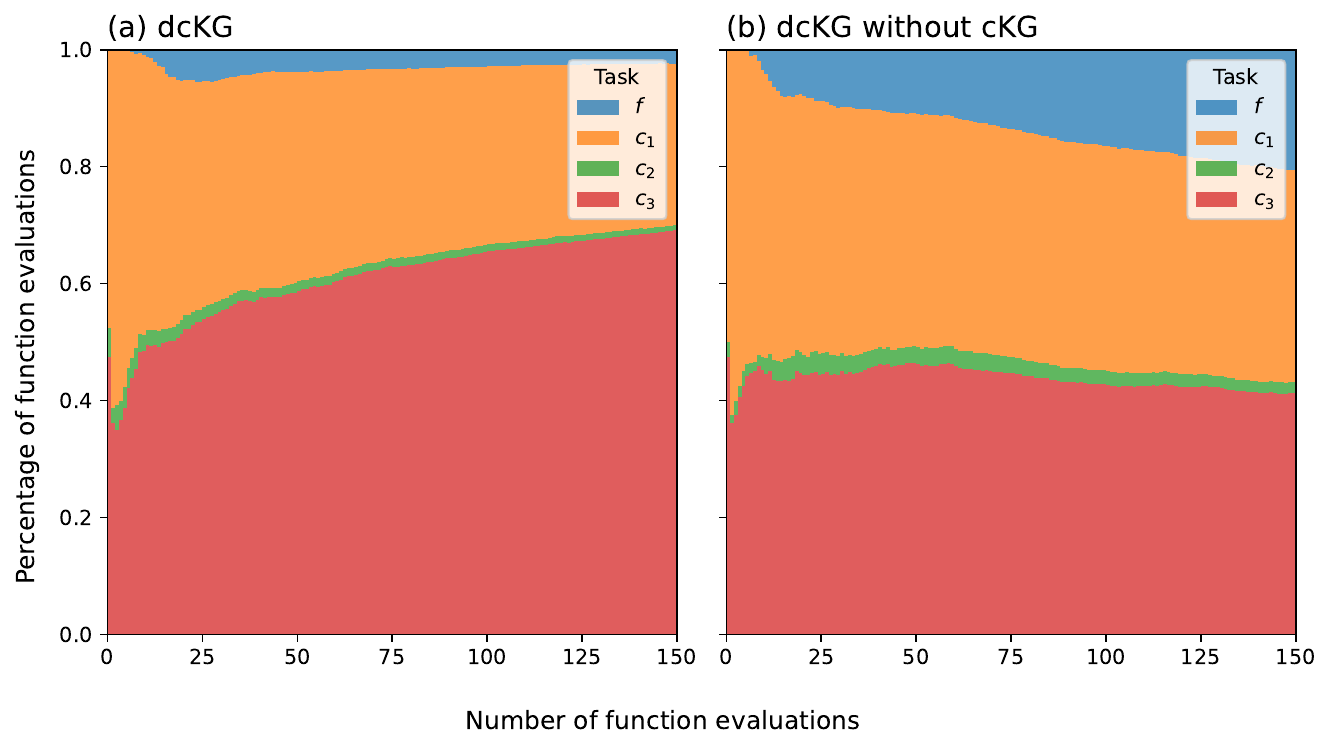}
    \caption{Test function: cumulative evaluations for each different function. \textbf{(a)} Using dcKG. \textbf{(b)} Using dcKG without cKG.}
    \label{fig:ablation_evals}
\end{figure}

Lastly, we conduct an ablation study on the proposed dcKG algorithm. In particular, we would like to see the performance with and without the inclusion of cKG as an additional acquisition function in deciding which objective/constraint to evaluate. We choose Test Function 2 for this study as it has multiple constraints, with one being inactive. The performance across 50 runs is shown in Fig.~\ref{fig:ablation}. The median value is observed to be comparable for the versions. However, it can be noted that for dcKG without cKG, the variance in performance is significantly larger. Thus, the performance of dcKG is much more reliable than the version without cKG. The cumulative evaluations for the two versions are shown in Fig.~\ref{fig:ablation_evals}. It can be seen, consistent with previous sections, that dcKG evaluates the constraints much more frequently than the objective. dcKG without cKG tends to evaluate objective values relatively more often, as well as evaluates the redundant constraint~($c_2$) more than dcKG. 

\begin{figure}
    \centering
    \includegraphics[width=0.7\linewidth]{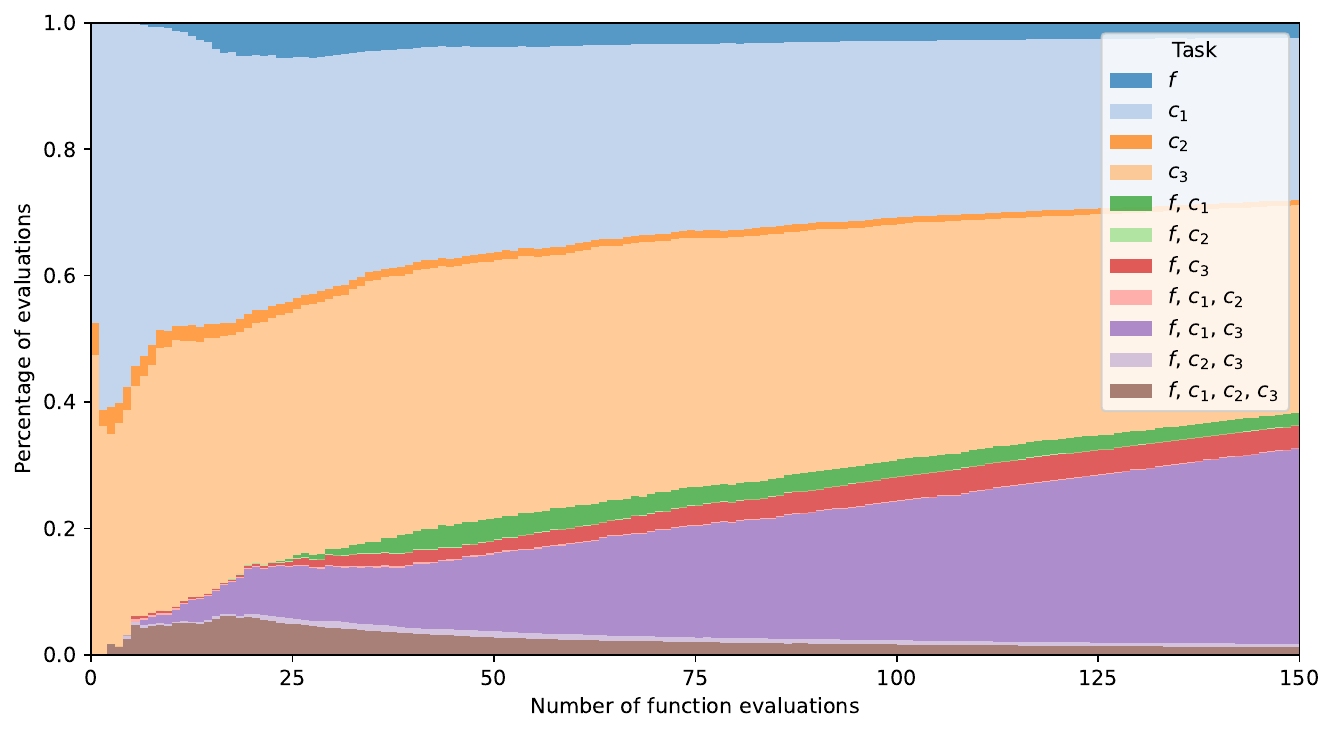}
    \caption{Test function: cumulative evaluations for each different task using dcKG}
    \label{fig:ablation_evals_fine}
\end{figure}

If we dig deeper into the evaluated tasks (Fig.~\ref{fig:ablation_evals_fine}), we see that later into the optimisation loop, dcKG starts to prefer coupled evaluations of $\{f, c_1\}$, $\{f, c_3\}$, $\{f, c_1, c_3\}$. These are evaluations of the objective function coupled with one or both of the two constrains bounding at the optimum. We believe these kinds of coupled evaluations allow dcKG to escape local maxima much faster than dcKG without cKG.

\section{Conclusion and Further Work}
\label{sec:con}

Many real-world optimization problems have multiple constraints. And often, they can be evaluated independently, and are not equally important. For example, not all constraints are binding near the optimum, not all are equally costly, and not all are equally difficult to learn. Thus, there is an obvious opportunity for Bayesian optimization algorithms to save computational resources by evaluating constraints selectively. Despite the apparent potential, there have been limited works in the domain of Bayesian Optimization that tackle problems with decoupled constraints. In this paper, we proposed two novel approaches based on the Knowledge Gradient acquisition function.

The main proposed approach, decoupled constrained Knowledge Gradient~(dcKG) extends cKG by evaluating the benefit of sampling each constraint individually, and supplements it further by using the coupled cKG criterion to avoid sampling bias towards well-known regions. The second approach, cEI+, is presented as a computationally faster alternative, whereby the sampling location is determined using constrained Expected Improvement, while the constraint to be evaluated is identified using dcKG at that point.   

Systematic numerical experiments were conducted on well-established benchmarks for various scenarios, including cases where objective and all constraints incur the same cost, and those where the costs are different. The experiments demonstrated that the decoupled methods perform better than coupled methods overall, and also outperformed the existing state-of-the-art method~(PESC). The performance difference from coupled approaches is particularly pronounced for problems which contained several redundant constraints, or if the redundant constraints were the more expensive ones. This was attributed to correct identification of binding constraints and efficient allocation of computational budget between the constraints and the objective.  

In future work, it should be straightforward to adapt our algorithms to the case where certain subsets of the objective/constraints are coupled, or to extend it to solve multi-objective optimization problems. Other relevant extensions include noisy optimization problems, and high-dimensional optimization problems.

\begin{acks}
We gratefully acknowledge support by the Engineering and Physical Sciences Research Council through the Mathematics of Systems II Centre for Doctoral Training at the University of Warwick (reference EP/S022244/1).
\end{acks}

\bibliographystyle{ACM-Reference-Format}
\bibliography{references, AR_refs}
%\usepackage{appendix}
\section{Appendix}
\appendix

\section{Parameter details}
For all experiments, the Bayesian optimization procedure was configured with the following parameters. To identify the best location according to the posterior mean, we used BoTorch's \texttt{optimize\_acqf} routine with a gradient-based and multi-start local optimization strategy. This step employed 20 random restarts and 2048 raw samples to initialize the search. For acquisition function optimization, the same routine was applied with 15 random restarts and 72 raw samples. In addition, since the acquisition function typically converges near the optimal location, we included the best posterior mean location as a smart initialization alongside the 72 raw samples. In both cases, optimization was performed using the L-BFGS-B algorithm from SciPy as the optimization routine for  \texttt{optimize\_acqf}.

For acquisition functions that rely on the Knowledge Gradient, we employ a deterministic quantile-based fantasy sampling scheme inspired by the Hybrid KG construction of \cite{Ungredda2022EfficientOptimization}. For cKG, we generate 7 deterministic Gaussian quantile fantasies for the objective. For each of these 7 objective fantasies, we generate 5 Sobol quasi-random K-dimensional vectors to represent the constraint fantasies. The resulting Cartesian product yields 7 x 5 = 35 deterministic posterior realizations per cKG evaluation. For each realization, the internal optimizer is run to obtain a candidate maximizer, and the overall cKG value is computed by averaging the improvements across all 35 fantasies.

For dcKG, fantasies are generated independently for each source. Each source employs a quantile sampler with 7 evenly spaced Gaussian quantiles, producing 7 deterministic posterior realizations per potential sampling source. Because dcKG conditions the model on a single scalar observation at a time, this one-dimensional quantile set fully captures the relevant posterior variability for each source. For both, cKG and dcKG, we employ Adam as an solver for the inner-optimization, using 15 restarts and 100 raw samples.

\section{Synthetic Test Functions}
\label{Appendix:tf}
\subsection{Mystery Function}
A two-dimensional problem with one constraint,
\[
\max_x f(x)=-2-0.01(x_2-x_1^2)^2 -(1-x_1)^2 - 2(2-x_2)^2 -7\sin(0.5x_1)\sin(0.7x_1x_2),
\]
subject to 
\[
c = -\sin\left(x_1-x_2-\frac{\pi}{8}\right) \leq 0,
\]
on the domain 
\[
x_1 \in [0,5], x_2 \in [0,5].
\]

\subsection{Constrained Branin Function}
A two-dimensional problem with one constraint,
\[
\max_x f(x)=(x_1-10)^2+(x_2-15)^2,
\]
subject to 
\[
c = \left(x_2 - \frac{5.1}{4\pi^2}x_1^2 + \frac{5}{\pi}x_1-6\right)^2+10\left(1-\frac{1}{8\pi}\right)\cos(x_1) + 5 \leq 0 
\]
on the domain 
\[
x_1 \in [-5,10], x_2 \in [0,15].
\]
\subsection{Test Function 2} A two-dimensional problem with three constraints, \[
\max_x f(x)=(x_1-1)^2 +(x_2-0.5)^2, 
\] subject to \[
c_1 = (x_1-3)^2 + (x_2+1)^2 -12 \leq 0,
\] \[
c_2 = 10x_1+x_2-7 \leq 0,
\]
\[
c_3 = (x_1-0.5)^2 + (x_2-0.5)^2 -0.2 \leq 0,
\]
on the domain 
\[
x_1 \in [0,1], x_2 \in [0,1].
\]

\section{Fitting a Gaussian Process}
\label{Appendix:GP}
We discuss how to fit an $n-$dimensional Gaussian Process when given the training dataset denoted by $D_f=\{ (x_i,y_i)\}_{i=1}^n$. Note that "fitting" a GP means finding its mean and variance.
Note that kernel (correlation structure) is prior, and we essentially impose a correlation between the objective function values at different points, so we need to use some prior understanding of the problem setting to give reasonable hyperparameters for the kernel function.
A typical choice is squared exponential kernel, which for two design vectors $\bar x, \bar x'$ takes the form:
\[
k(\bar x, \bar x') = \sigma_f^2 \exp\left(-\sum_{i=1}^n \frac{1}{2} \frac{x_i - x_i'}{l_i^2}\right) + \sigma_{noise}^2 \delta(\bar x, \bar x'),
\]
where we here use the notation $m(x)$ for the mean and $k(x,x')$ for the covariance.

Here $\sigma_f, \sigma_{noise}, l_i$ are hyperparameters that determine the smoothness of the functions. In practice, we estimate them via maximum likelihood. Note that even when dealing with noiseless data, it is often useful to suppose in the model that there is tiny noise, e.g. $10^{-4}$. This often improves numerical stability during inference.

Suppose $f$ is a GP, then the vector 
\[
\left( f(x_1), \dots, f(x_n), f(x) \right) \sim N_{n+1}(0, \Sigma),
\]
where
\[
\Sigma= 
\begin{pmatrix}
k(x_1,x_1) & \dots & k(x_1, x_n) & k(x_1,x)\\
\dots \\
k(x_n,x_1) & \dots & k(x_n, x_n) & k(x_n,x)\\
k(x,x_1) & \dots & k(x, x_n) & k(x,x)
\end{pmatrix}
=
\begin{pmatrix}
k_{XX} & k_X(x, \bar x)\\
k_X(x, \bar x)^T & k(x,x)\\
\end{pmatrix}.
\]

Here we use notation:
\[
X=(x_1, \dots, x_n),
\]
\[
K_{XX}=
\begin{pmatrix}
k(x_1,x_1) & \dots & k(x_1, x_n)\\
\dots \\
k(x_n,x_1) & \dots & k(x_n, x_n)
\end{pmatrix},
\]
\[
k_X(x, \bar x)=(k(x_1,x), \dots, k(x_n,x))^T,
\]
\[
F=(f(x_1), \dots, f(x_n))^T.
\]

Then we can derive through certain properties of Gaussians that the exact expression for the conditional distribution is:
\[
f(x) \vert (f(x_1), \dots, f(x_n)) \sim N(\bar \mu(x, \bar x), \bar k(x, \bar x)),
\]
where 
\[
\bar \mu(x, \bar x) = k_X(x, \bar x)^T K_{XX}^{-1} F,
\]
\[
\bar k(x, \bar x) = k(x,x) -k_X(x,\bar x)^T K_{XX}^{-1} k_X(x,\bar x).
\]
In the case of noisy observations, that is $y_i=f(x_i) + N(0,\sigma^2)$, the mutual distribution is the following:
\[
\left( f(x_1), \dots, f(x_n), f(x) \right) \sim N_{n+1}(0, \Sigma),
\]
with 
\[
\Sigma=
\begin{pmatrix}
k_{XX} +\sigma^2I & k_X(x, \bar x)\\
k_X(x, \bar x)^T & k(x,x)\\
\end{pmatrix}.
\]
Hence the conditional distribution is given by,
\[
f(x) \vert (f(x_1), \dots, f(x_n)) \sim N(\bar \mu(x, \bar x), \bar k(x, \bar x)),
\]
where 
\[
\bar \mu(x, \bar x) = k_X(x, \bar x)^T (K_{XX} +\sigma^2I)^{-1} F,
\]
\[
\bar k(x, \bar x) = k(x,x) -k_X(x,\bar x)^T (K_{XX}+\sigma^2I)^{-1} k_X(x,\bar x).
\]

\section{Contextualisation of Work in ICLR paper \cite{Nguyen2024OptimisticConstraints}}

In our context, the problem at hand can be described as:

\begin{align}
    \max_{x\in\mathcal{X}} f(x) ~|~ g(x) \leq 0 \forall g \in \mathcal{G},
\end{align}

where, $f(\cdot)$ is the objective function, $g(\cdot)$ is a constraint function from the set of constraints $\mathcal{G}$ with $x\in\mathcal{X}$ being the decision vector. This requires reframing of the definitions presented in \cite{Nguyen2024OptimisticConstraints} for a faithful implementation. Below we discuss how the definitions were adapted for our context.

We define functions for upper and lower confidence bounds $u_{h, t-1}(\cdot)$ and $l_{h, t-1}(\cdot)$ for a function $h(\cdot)$ with a Gaussian process (GP) model trained on data collected until $(t-1)$ iterations as follows:
\begin{align}
    u_{h, t-1} (x) = \mu_{h, t-1}(x) + \left(\beta \times \sigma^2_{h, t-1}(x)\right)^{\frac{1}{2}},\\
    l_{h, t-1} (x) = \mu_{h, t-1}(x) - \left(\beta \times \sigma^2_{h, t-1}(x)\right)^{\frac{1}{2}},
\end{align}
where $\mu_{h, t-1}(x)$ and $\sigma^2_{h, t-1}(x)$ are the predictive mean and variance from the GP model of $h(x)$, and $\beta = 2 \log \left[\frac{(|\mathcal{G}|+1) |\mathcal{X}| t^2 \pi^2}{6\delta}\right]$ with $\delta=0.1$.

The vertical exploration bound is defined as: $v_t = 2 \left(\beta \times \sigma^2_{f, t-1}(x)\right)^{\frac{1}{2}}$, based on the predictive variance for the objective function.

With these, the optimistic feasible set of solutions is defined as:
\begin{align}
    O_t = \{x \in \mathcal{X} ~|~ l_{g, t-1}(x \leq 0) \}.
\end{align}

Naturally, the complement set, i.e. the set of infeasible solutions, is given by $O'_t = \{x \in \mathcal{X} ~|~ l_{g, t-1}(x > 0) \}$.

Furthermore, given the cost of evaluation for a function $h(\cdot)$ is $c_h >0$, the uncharted region due to $v_t$-relaxed feasible confidence region is defined as $U_t = \{x \in \mathcal{X} ~|~ u_{g, t-1}(x) > \frac{v_t}{c_f} \}$. Thus, the $v_t$-relaxed feasible confidence region is defined as $S_t = O_t \setminus U_t$.

Following these, the acquisition function becomes:
\begin{align}
    \alpha(x) = \begin{cases}
u_{f, t-1}(x) & \text{if } x \in O_t \\
\rho & \text{otherwise},
\end{cases}
\end{align}
where, $\rho << 0$ is a penalisation constant, and we seek $x^* = \arg \max_{x \in \mathcal{X}} \alpha(x)$.

We will evaluate the objective function if $x^* \in S_t$. In case, $x^* \in U_t$, we will evaluate the $k$th constraint that violates the feasibility most as determined by:
\begin{align}
    k = \arg\max_{g\in\mathcal{G}} \frac{u_{g,t-1}(x^*)}{c_g}.
\end{align}

\end{document}